\documentclass{article}
\usepackage[font={small}]{caption}
\usepackage{amsmath}
\usepackage{cite}
\usepackage{arxiv}
\usepackage{graphicx}
\usepackage[utf8]{inputenc} 
\usepackage[T1]{fontenc}    
\usepackage{hyperref}       
\usepackage{url}            
\usepackage{booktabs}       
\usepackage{amsfonts}       
\usepackage{nicefrac}       
\usepackage{microtype}      
\usepackage{lipsum}
\usepackage{float}
\usepackage{subfig}

\usepackage{authblk}
\usepackage{xcolor}

\title{Bibliography management: \texttt{natbib} package}
\title{A Brief Review of Deep Multi-task Learning and Auxiliary Task Learning}
\date{}
\author{
  
  Partoo Vafaeikia, Khashayar Namdar, Farzad Khalvati \\
  Department of Medical Imaging\\
  University of Toronto \\
  Toronto, ON, Canada\\
  \texttt{partoo.vafaeikia@mail.utoronto.ca} \\
  \texttt{ernest.namdar@utoronto.ca} \\
  \texttt{farzad.khalvati@utoronto.ca} 
  
}
\begin{document}

\maketitle

\begin{abstract}
Multi-task learning (MTL) optimizes several learning tasks simultaneously and leverages their shared information to improve generalization and the prediction of the model for each task. Auxiliary tasks can be added to the main task to ultimately boost the performance. In this paper, we provide a brief review on the recent deep multi-task learning (dMTL) approaches followed by methods on selecting useful auxiliary tasks that can be used in dMTL to improve the performance of the model for the main task.
\end{abstract}
\section*{Introduction}
Multi-task learning (MTL) is broadly used across various applications of machine learning and has several advantages in comparison with the single-task learning. Since layers are shared between different tasks and features are not repeatedly calculated for each task, the amount of memory used is reduced and the inference speed is improved. In addition, if tasks share complimentary information, they act as regularizers for each other which results in the improvement of the prediction performance of each task \cite{v2020revisiting}. This has been proven in various areas such as detection and classification \cite{ren2015faster}, computer vision \cite{misra2016crossstitch,kokkinos2016ubernet}, depth estimation \cite{Eigen_2015_ICCV}, natural language processing \cite{Collobert:2008:UAN:1390156.1390177,conf-coling-2018,mccann2018natural} and drug discovery \cite{ramsundar2015massively}. The goal of this review paper is to provide an overview of various deep multi-task learning (dMTL) solutions and possible improvements in performance through efficient auxiliary tasks selection. One of the main challenges in dMTL is to decide what to share across tasks to achieve better results by avoiding negative transfer, which describes a situation where an improper feature sharing scheme would negatively impact the overall performance of the predictions instead of improving it. In earlier approaches, this has mostly been done through hand-designed architectures where the network is manually tuned to achieve the best performance. However, there is a large number of parameters for alteration, specially in deep MTL, that makes it impossible to find the best architecture with this approach. Therefore, in the first part of this review, we discuss dMTL architectures that find the most optimal setting to tackle this issue. In addition to the network architecture, a successful utilization of auxiliary tasks is of great importance for improving the performance of the model for the main task~\cite{du2018adapting}. Therefore, in the second part of this review, we provide a few methods that could be used to decide which auxiliary tasks should be trained along with the main task.

\section*{Multi-Task learning for deep learning }
\begin{figure}[ht]
\includegraphics[width=0.95\textwidth]{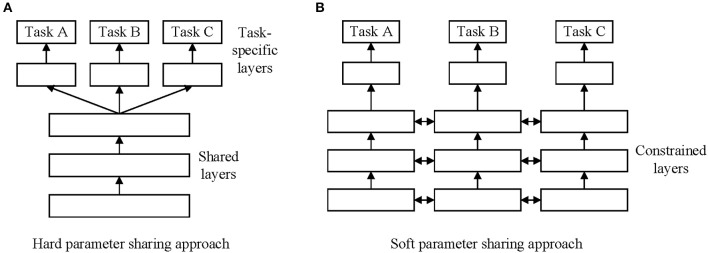}
\centering
\caption{ (A) describes hard parameter sharing where hidden layers are shared among all tasks and (B) shows a soft parameter sharing with each task having its own hidden layers. Reprinted from \cite{10.3389/fnins.2019.01387}.}
\label{h&S}
\centering
\end{figure}

Classic approaches in dMTL are generally categorized into two main architectures: hard \cite{Caruana93multitasklearning:} and soft  \cite{duong-etal-2015-low} parameter sharing (Figure~\ref{h&S}). In hard parameter sharing, hidden layers are generally shared among all tasks and there are only a few task specific output layers. Hard parameter sharing is more commonly used in dMTL and it has a low risk of overfitting since the tasks are learned simultaneously which results in improved generalization \cite{Baxter97abayesian/information}. However, hard parameter sharing performs well only if tasks are closely related, therefore, new approaches have focused more on learning what features to share between tasks. In soft parameter sharing, hidden layers and parameters are task specific and have their own settings. In addition, parameters are regularized in order to encourage them to be similar \cite{ruder2017overview}. Shared layers learn the intrinsic features of the data, while task specific layers classify data into output layers using the learned latent features of previous layers \cite{Thung2018ABR}. Recent works in dMTL have attempted to improve the soft and hard parameter sharing methods and build more optimized mechanisms, which will be discussed in the following sections.

\subsection*{Multilinear Relationship Networks}
\begin{figure}[ht]
\includegraphics[width=0.75\textwidth]{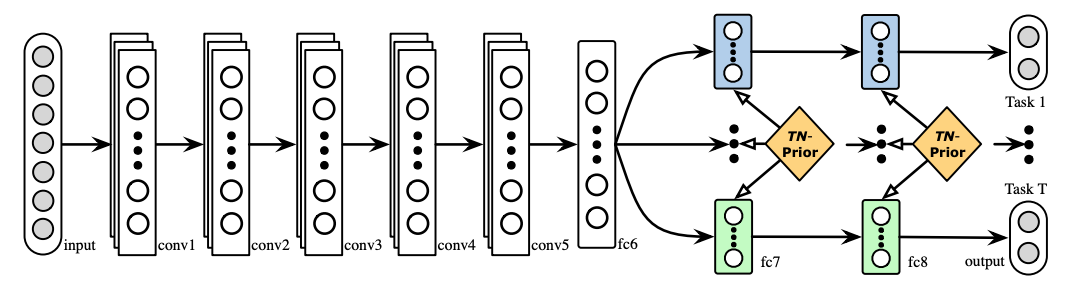}
\centering
\caption{Multilinear Relationship Networks (MRNs). $fc7-fc8$ fully-connected layers are task specific layers and tensor normal (TN)-prior units are tensor normal priors on multiple task-specific
layers to learn tasks relationships. Reprinted from \cite{long2015learning}.}
\label{MRN}
\centering
\end{figure}

One area where traditional dMTL models can be enhanced is to improve feature transferability and utilization of initial layers’ features in task-specific layers. Multilinear Relationship Networks (MRNs) \cite{long2015learning} start with convolutional neural network (CNN) and a fully connected layer to learn shared features among tasks and continue to have separate stacks of fully connected layers for individual tasks (Figure~\ref{MRN}). Tensor priors model tasks' parameters to learn multilinear relationships between the features, classes and tasks. These priors are added to the fully connected layers to help transferring knowledge across different tasks. This approach tries to mitigate both negative transfer and under-transfer problems. However, as a result of its predefined architecture, the network can only be used for specific tasks effectively \cite{ruder2017overview}.

\subsection*{Cross-stitch Networks}
\begin{figure}[h]
\includegraphics[width=0.55\textwidth]{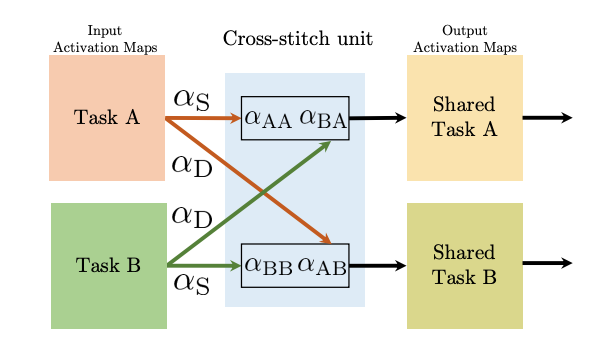}
\centering
\caption{Cross-stitch units learn a linear combination of the previous layer outputs and pass it to next layers of the network. This linear combination is parameterized using $\alpha$. $\alpha_{D}$ are different-task values and $\alpha_{S}$ are similar-task values. Reprinted from~\cite{misra2016crossstitch}.}
\label{CSU}
\centering
\end{figure}

To improve the generalization of previous methods, cross-stitch networks proposed separate network architectures for each task similar to soft parameter sharing~\cite{misra2016crossstitch}. Cross-stitch units are additions to this architecture which combine multiple task-specific networks into one model (Figure~\ref{CSU}). This can balance tasks’ shared features as well as task-specific representations. Cross-stitch units learn a linear combination of the activation maps and pass it to the next layers. What is not covered in this method is the selection process for tasks to be learned together efficiently, in order to maximize performance. Cross-stitch networks only determine the best way to leverage features of given tasks. Therefore, it is the responsibility of the algorithm designer to choose tasks efficiently before passing them to this network. Furthermore, it is not clear where these units need to be inserted for maximum performance \cite{misra2016crossstitch}.

\subsection*{Using Uncertainty to Weigh Losses for Scene Geometry and Semantics}
\begin{figure}[ht]
\includegraphics[width=0.75\textwidth]{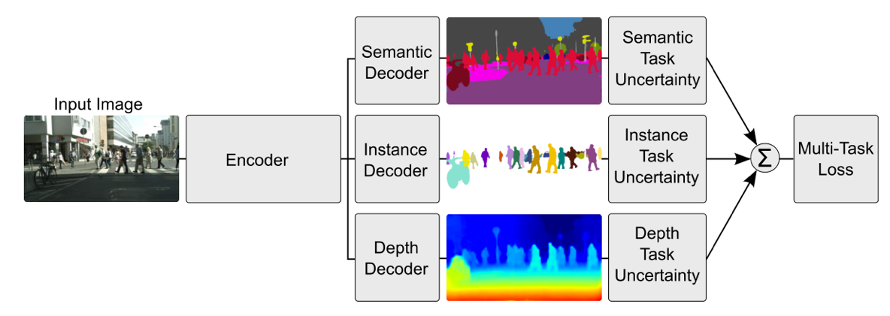}
\centering
\caption{High level summary of deep convolutional
encoder-decoder network architecture that incorporates tasks uncertainties to build multi-task loss on CityScapes dataset. Reprinted from \cite{kendall2017multitask}.}
\label{UW}
\centering
\end{figure}

This approach utilizes the homoscedastic uncertainty of tasks which is a task-dependent uncertainty that captures the relative confidence between tasks, as a basis to weighing losses and adjust their relative weights in the model’s cost function as opposed to tuning the weights manually. In other words, instead of manipulating internal parameters of the network, the cost function is dynamically aligned during the training phase. This allows the network to simultaneously learn multiple objectives and allows weights to be dynamic during training. A unified network with this method is used to simultaneously learn multiple objectives and efficiently weigh them for semantic segmentation, instance segmentation and depth regression. Kendall et al. \cite{kendall2017multitask} developed a model using this method on CityScapes \cite{cordts2016cityscapes}, a dataset for road scene understanding. This model takes in an RGB image and simultaneously produces a pixel-wise classification, semantic segmentation and pixel-wise depth estimation. This model is shown in Figure \ref{UW} and consists of several convolutional encoders to produce a shared representation, as well as multiple decoders for each task specifically \cite{kendall2017multitask}.

\subsection*{Sluice Networks}
\begin{figure}[ht]
\includegraphics[width=0.5\textwidth]{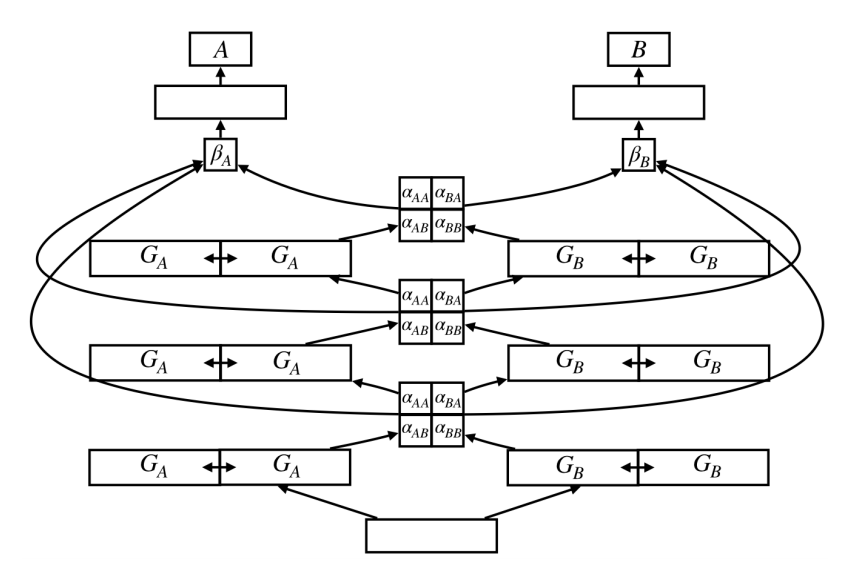}
\centering
\caption{Sluice network that consists of one shared input layer, three hidden layers for each task and two task specific output layers. $\alpha$ parameter decides which features to be shared between the two task-specific networks and $\beta$ controls which final outputs layers to be used for prediction. Reprinted from \cite{Ruder2017SluiceNL}.}
\label{SN}
\centering
\end{figure}

The last network that we discuss in this review is Sluice network which generalizes some of the methods we reviewed earlier such as hard parameter sharing and cross-stitch networks~\cite{Ruder2017SluiceNL}. This architecture learns layers and subspaces to be shared and is capable of using specific parts of layers for different tasks. Skip connections are implemented to control which internal blocks feed the ultimate task-specific outputs (Figure~\ref{SN}). Sluice network has shown to outperform some of previous networks as a result of a more comprehensive and general approach to dMTL. However, in this method, features are shared with equal weights and unhelpful tasks are of the same importance as the useful ones. This may increase the possibility of misleading predictions.

\section*{Recent Approaches on Auxiliary Task Selection}
As discussed earlier, auxiliary tasks are influential in MTL as they help the model improve and generalize better for the main task. However, if not selected properly, they can negatively impact MTL models \cite{guo2019autosem}. Therefore, choosing the right auxiliary tasks in multi-task learning is crucial. This topic has received significant attention in machine learning in recent years and various approaches have been proposed on how to select optimal auxiliary tasks for a given main task. In all these approaches, the optimal auxiliary tasks are selected from a pool of predefined tasks.

\subsection*{Gradient Similarity for Auxiliary Losses}
Du et al. proposed that there should be a similarity between the main tasks and the auxiliary tasks for the MTL to be successful and to be preferred over a single task learning approach~\cite{du2018adapting}. Their proposed method measures the similarities between the main and auxiliary tasks using cosine similarity. This gives a measure to understand which auxiliary tasks are helpful in the training. The ultimate goal is to utilize the helpful auxiliary losses and block negative transfer where they are nonconstructive. Training on an auxiliary task is chosen only if its gradient has non-negative cosine similarity with that of the main task. This method is not domain specific and can be utilized in various areas. However, cosine similarity does not guarantee the usefulness of tasks and only removes those that certainly have negative impact on the learning process.

\subsection*{Auxiliary Modules in dMTL}
\begin{figure}[ht]
\includegraphics[width=0.7\textwidth]{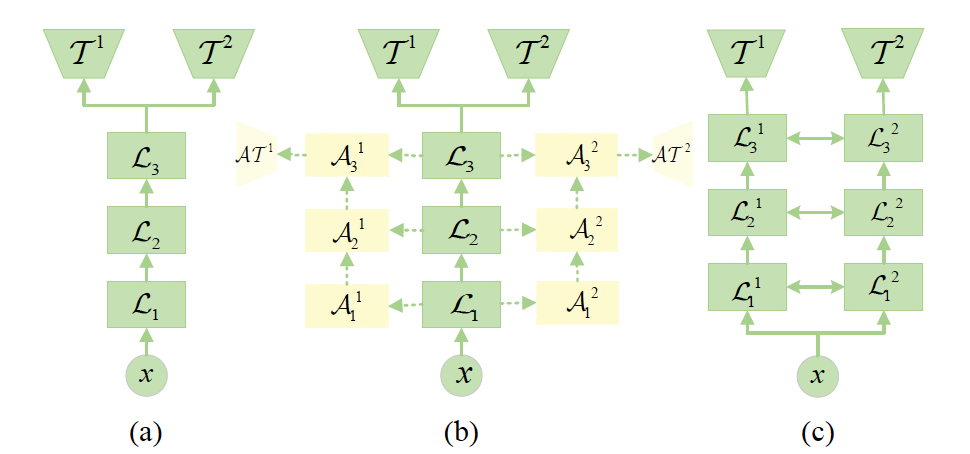}
\centering
\caption{A comparison between hard parameter sharing (a), soft parameter sharing (c), and the proposed network (b). The auxiliary module $A^t_l$ acts as a regularization for inductive transfer in layer $l$ and task $t$. This module mimics the characteristics of the soft parameter sharing. Reprinted from \cite{liu2019auxiliary}.}
\label{AM}
\centering
\end{figure}

This approach tries to combine the two original methods, hard and soft parameter sharing, into one model and integrate the benefits of the soft parameter sharing to assist the training of a hard sharing network. The main network and the auxiliary modules are optimized together during training where the auxiliary modules serve as a regularization by introducing inductive bias to create a balance between the shared and task-specific representations in hidden layers. These additional modules are removed during testing phase to avoid adding extra computational burden on the network (Figure~\ref{AM}) \cite{liu2019auxiliary}.

\subsection*{Gated Multi-task Network}
\begin{figure}[ht]
\includegraphics[width=0.6\textwidth]{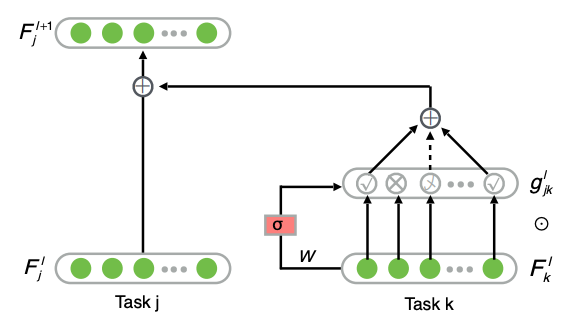}
\centering
\caption{Gated Multi-Task Network: When task $j$ borrows the features
from task $k$, gate $g$ is inserted to select the helpful features. The gate is calculated using $\sigma$, the scalar weight between two tasks that reflects the degree of association between them. Reprinted from \cite{Xiao2018GatedMN}.}
\label{GMTN}
\centering
\end{figure}

This network proposed the gated sharing units which filter the flow of features between different task-specific layers to reduce interference~\cite{Xiao2018GatedMN}. These units act like a gate mechanism which only allows the helpful auxiliary tasks to share features and evidence with the main task. While every task has its own network, the gated sharing units selectively allow features to be exchanged across task-specific layers and get merged. This is done through inserting a scalar weight between every two tasks. This weight is updated by back-propagation and shows the similarity between the two tasks (Figure~\ref{GMTN}). The advantage of this method is that it selects and calculates appropriate weights for auxiliary tasks simultaneously which not only prevents harmful tasks from exchanging features with the main task but also allows the most useful tasks to be more impactful by allocating higher weights to them. However, in Li et al. experiments, this method has shown unstable results on different models \cite{li2019empirical}.

\subsection*{AUTOSEM: Automatic Task Selection and Mixing in Multi-Task Learning}
\begin{figure}[ht]
\includegraphics[width=0.95\textwidth]{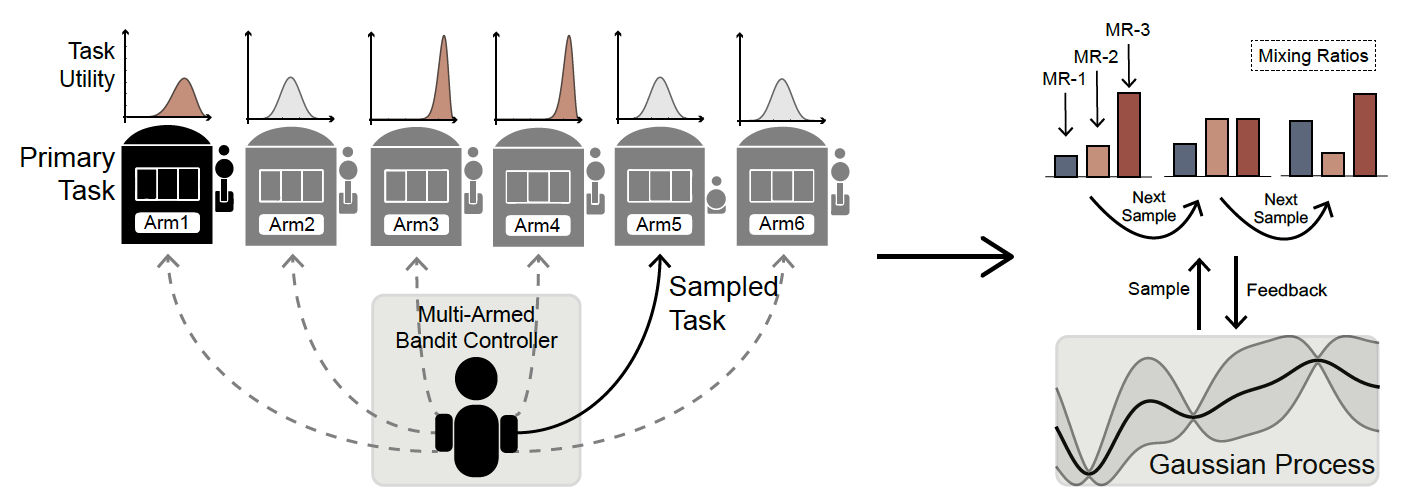}
\centering
\caption{An overview of AUTOSEM framework. Left is a demonstration of the multi-armed bandit controller used for task selection and on right a learning process for automatically finding the mixing ratio is shown. Reprinted from \cite{guo2019autosem}.}
\label{AS}
\centering
\end{figure}

Selecting best auxiliary tasks can be done through hyper-parameter tuning over all possible combinations or manual intuition. However, this approach is error prone and human bias might be a barrier. Autosem framework divides this optimization into two parts. First stage is automatic task selection of auxiliary tasks from all possible options. A non-stationary multi-armed bandit controller is utilized which dynamically alternates between auxiliary tasks choices and returns an estimate on their usefulness with respect to the main task. The second stage automatically learns a training mixing ratio of the selected auxiliary tasks using a Bayesian optimization framework (Figure~\ref{AS}). The performance of each setup is modeled as a Gaussian Process to find the optimal mixing ratio value \cite{guo2019autosem}. In contrast to previous methods, this approach has two stages that each addresses a different problem. They show that having both stages improve the performance of the model compared to a case where one stage is excluded and that they are both crucial in the process.

\section*{Conclusion}
In this paper, we have briefly discussed the motivation for deep multi-task leaning (dMTL) and the benefits of having auxiliary tasks in dMTL. In addition, we discussed recent methods on dMTL as well as auxiliary task selection. A combination of the two techniques creates a framework that maximizes model performance. 

\section*{Acknowledgement}
This research has been supported by Huawei Technologies Canada Co., Ltd.
\bibliography{ref} 

\bibliographystyle{ieeetr}

\end{document}